# Classification Of Gradient Change Features Using MLP For Handwritten Character Recognition

Sandhya Arora[1] , Latesh Malik[2] , Debotosh Bhattacharjee[2] , Mita Nasipuri[3]

[1] Deptt. Of Comp. Sc. & En gg., Meghnad Saha Institute of Technology, Kolkata, sandhyabhagat@yahoo.com

[2] Deptt. Of Comp. Sc. & Engg., G.H. Raisoni College of Engineering, Nagpur,University Of Calcutta, Kolkata, lateshmalik@rediffmail.com , debotoshb@hotmail.co m

[3] Deptartment Of Compter Science & Engineering,Jadavpur University,Kolkata, mitanasipuri@yahoo.com

*Abstract:* - A novel, generic scheme for off-line handwritten English alphabets character images is proposed. The advantage of the technique is that it can be applied in a generic manner to different applications and is expected to perform better in uncertain and noisy environments. The recognition scheme is using a multilayer perceptron(MLP) neural networks. The system was trained and tested on a database of 300 samples of handwritten characters. For improved generalization and to avoid overtraining, the whole available dataset has been divided into two subsets: training set and test set. We achieved 99.10% and 94.15% correct recognition rates on training and test sets respectively. The purposed scheme is robust with respect to various writing styles and size as well as presence of considerable noise.

## I. Introduction

Optical Character Recognition (OCR) is a process of automatic computer recognition of characters in optically scanned and digitized pages of text. OCR is one of the most fascinating and challenging areas of pattern recognition with various practical application potentials. It can contribute immensely o the advancement of an automation process and can improve the interface between man and machine in many applications. Some practical application potentials of OCR system are: 1) reading aid for the blind 2) automatic text entry into the computer for desktop publication, library cataloging, ledgering etc 3) automatic reading for sorting of postal mail, bank cheques and other documents 4) document data compression: from document image to ASCII format 5) language processing 6) multi-media system design etc.

Intensive research has been done on optical character recognition (OCR) and many commercial OCR systems are now available in the market but most of these system work for Roman, Chinese, Japanese, and Arabic characters. There is no sufficient work on Indian language character recognition although there are 12 major scripts in India.

Numerous techniques for offline handwritten recognition have been investigated based on direct matching, relaxation matching, Discriminant Analysis for Arabic characters[, Hidden Markov Models for English words, Hough transform technique for Chinese character, Bayesian classifier for printed characters, Support vector machines , prototype matching for multifont characters, and local Affine transform for handwritten numerals etc.

## II. The Approach

The overall approach is described as below. Initially the character image is extracted from background and converted to binary format. Then some preprocessing is performed and character image is divided into different segments. Feature vector is constructed from these segments and fed as an input to the Neural Network for recognition.

### A  Extraction of Character from image

The character image is extracted from the whole image by taking leftmost, rightmost, bottommost, topmost black pixel positions of the character. Then the character is kept in a new image with the height and width found in the above step.

### B  Preprocessing

As preprocessing, we considered only size normalization. The 16 level gray scale image obtained is initially converted into binary image by assigning every pixel value equal or greater than the threshold value, by value 1 and all other a value 0. This binary image is scaled and thinned to 100 X 100 image. No further preprocessing like tilt correction, smoothing etc are considered. The threshold value we are considering here an integer value 7000000 which is representing the gray level intensity value of a pixel.

### C   Scaling of extracted image

We are using Affine Transformation to perform a linear mapping from 2D coordinates to other 2D coordinates that preserves the "straightness" and 'parallelness" of lines. Affine transformation can be constructed using sequences of translations, scales, flips, rotations and shears. Image is scaled in 100x100 pixel resolution.

[x'] [m00 m01 m02][x]
[y']=[m10 m11 m12][y]
[1 ] [ 0  0  1 ][1]

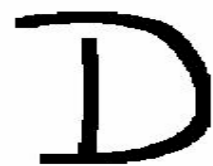 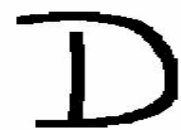

1. Input Image    2. Scaled Image

**Fig 1** Input character Image and scaled Ima ge

### D   Thinning of scaled image

The thinning algorithm transforms an object to a set of simple digital arcs. The structure obtained is not influenced by small contour inflections that may be present on the initial contour. The basic approach is to delete from object's border points that have more than one neighbour in the object and whose deletions does not locally disconnect the object. Here a connected region is one in which any two points in the region can be connected by a curve that lies entirely in the region. In this way, end points of thin arcs are not deleted.

Let ZO(P1) count be the number of zeros to nonzero transitions in the ordered set P2,P3,P4,P5,P6,P7,P8,P9,P2. Let Nzcount(P1) be the number of non zero neighbours of P1.

```
P3  P2  P9
P4  P1  P8
P5  P6  P7
```

Then P1 is deleted if
Step 1: 2<=Nzcount<=6
Step 2: and ZO(P1)=1
Step 3: and P2.P4.P8=0 or ZO(P2) != 1
Step 4: and P2.P4.P6=0 or ZO(P4) != 1

The procedure[7], all above said steps are repeated until no further changes occur in the image. The corner positions are the special cases to be considered and taken care for more perfection in the thinning procedure

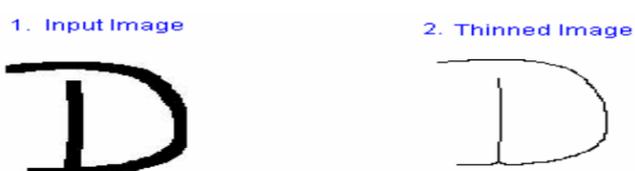

Fig 2. Input Scaled Image and Thinned Image

E Gradient change of thinned image edge & Analysis

Then the source image is skeletonized and divided into different segments(9,16,25 segments). For each segment perform the following steps[1,4]:-

Step 1: Consider two successive rows at a time. the whole process should be performed in horizontal manner.
Step 2: Let $B_i$ be the first black pixel found in row i
Step 3: Calculate the distance between $B_i$ and $B_i+1$
Step 4: Sum up the distances(gc's) for whole segment

These gc's values may be zero, negative, positive (Fig 3). It may contribute zero for horizontal or vertical segements (symmetrical around the x or y axis), may contribute positive for convex shapes, negative for concave shapes. The features(gc values) were extracted for each segment and the complete image was described with a component vector V=(gc1,gc2…..gc9) or V=(gc1,gc2….gc16) or V=(gc1,gc2….gc25).

The idea is that, in practice although for two different images, a few of the segments may yield identical gc values, it would be rare if all the components in the vector were same for completely different shapes.. The patterns obtained were used as inputs to the Neural Network for recognizing different characters.

Fig 3 gc Computation

### III. Neural Network Architecture

Use of ANN in handwritten recognition task has become very popular because of ANN tools(say MLP classifiers) perform efficiently when input data are often affected by noise and distortions. Also, the parallel architecture of a connectionist network model and its adaptive learning capability are added advantages. In out approach, we feed the feature vector of length 9,16,25 to MLP classifier.

During simulation we considered MLP's with one hidden Layer. For feature vector of size say 9 we have used 9x9x10 architecture, altogether there are 28 neurons. The training was therefore fast. The network was trained with the normalized data using conjugate-gradient( CG) method of training. This method was preferred to the gradient descent method, since CG takes into account the non-linearity of the surface. The CG procedure does not ask user to specify any parameters such as learning rate.

The basic back propagation algorithm adjusts the weights in the steepest descent direction (negative of the gradient). This is the direction in which the performance function is decreasing most rapidly. It turns out that, although the function decreases most rapidly along the negative of the gradient, this does not necessarily produce the fastest convergence. In the conjugate gradient algorithms a search is performed along conjugate directions, which produces generally faster convergence than steepest descent directions.

All of the conjugate gradient algorithms start out by searching in the steepest descent direction (negative of the gradient) on the first iteration.

$$p_0 = -g_0 \quad \text{------------------} \quad (1)$$

A line search is then performed to determine the optimal distance to move along the current search direction:

$$x_{k+1} = x_k + a_k p_k \quad (2)$$

Then the next search direction is determined so that it is conjugate to previous search directions. The general procedure for determining the new search direction is to combine the new steepest descent direction with the previous search direction:

$$p_k = -g_k + \beta_k p_{k-1} \quad (3)$$

where $x_k$ is a vector of current weights and biases, $g_k$ is the current gradient, and $a_k$ is the learning rate.

## IV. Experimental Results

Different classifiers have been used for handwritten digit recognition, such as statistical[3], structural and neural networks[5], 99.77% recognition rate have been reported for handwritten digits using multiple features & multiple neural networks[2] but for characters 92.3% recognition rate have been reported [6]. We can say that handwritten character recognition is still an open problem.

We have simulated the present recognition schema on 10 characters database of handwritten characters. This database includes 250 sample set for training set, 50 sample set for testing. A few samples are shown in Table 2. Ideal samples are shown in Table 1.

F L O Z

**Table 1** Ideal samples of alphabet

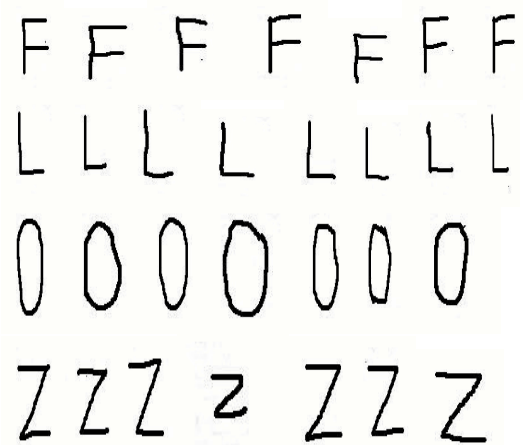

**Table 2** A typical sample data subsets of handwritten characters

We divided character image into different size segments. For each segment we computes gradient changes. These gradient values are normalized for training and testing of Neural Network. For different size segment (input vector to Neural Network), we have made several simulation runs varying the normalization factor and we have observed that recognition accuracy on test set of samples can be improved by taking optimal segment size and it can also be slightly improved using proper normalization factor. In table 3 we present these recognition results on both the training set and test set.

| Input Feature Vector Size: 9 | | |
|---|---|---|
| *Recognition Accuracy* | | |
| *Normalization Factor* | *Training Set* | *Test Set* |
| +(35/70) | 98.25 | 93.39 |
| +(40/80) | 97.28 | 90.15 |
| **Input Feature Vector Size: 16** | | |
| *Recognition Accuracy* | | |
| *Normalization Factor* | *Training Set* | *Test Set* |
| +(30/60) | 99.10 | 94.15 |
| +(40/80) | 98.75 | 93.89 |
| **Input Feature Vector Size: 25** | | |
| *Recognition Accuracy* | | |
| *Normalization Factor* | *Training Set* | *Test Set* |
| +(25/50) | 90.34 | 89.42 |
| +(30/60) | 90.10 | 86.10 |

**Table 3** Recognition Result